\documentclass[10pt,twocolumn,letterpaper]{article}

\usepackage[pagenumbers]{wacv} % To force page numbers, e.g. for an arXiv version

\usepackage{graphicx}
\usepackage{amssymb}
\usepackage{booktabs}
\usepackage{xspace}

\usepackage{amsmath,amsfonts}
\usepackage{algorithmic}
\usepackage{algorithm}
\usepackage{array}
\usepackage{textcomp}
\usepackage{stfloats}
\usepackage{url}
\usepackage{verbatim}
\usepackage{cite}
\usepackage{multirow}

\usepackage{color}

\definecolor{red}{rgb}{1.00,0.00,0.00}
\definecolor{blue}{rgb}{0.00,0.00,1.00}
\newcommand{\best}[1] {\textcolor{red}{#1}}
\newcommand{\second}[1] {\textcolor{blue}{#1}}

\usepackage[pagebackref,breaklinks,colorlinks]{hyperref}

\usepackage[capitalize]{cleveref}
\crefname{section}{Sec.}{Secs.}
\Crefname{section}{Section}{Sections}
\Crefname{table}{Table}{Tables}
\crefname{table}{Tab.}{Tabs.}

\def\wacvPaperID{1971} % *** Enter the WACV Paper ID here
\def\confName{WACV}
\def\confYear{2025}

\begin{document}

\title{EDMB: Edge Detector with Mamba}

\author{
Yachuan Li$^{1}$\thanks{This work is supported by National key r\&d program (Grant no. 2019YFF0301800), National Natural Science Foundation of China (Grant no. 61379106), Shandong Provincial Natural Science Foundation (Grant nos.ZR2013FM036, ZR2015FM011). China Scholarship Council (CSC) (Grant no. 202406450075).}
\qquad
Xavier Soria Poma$^{2}$
\qquad
Yun Bai$^{1}$
\qquad
Qian Xiao$^{1}$\\
\qquad
Chaozhi	Yang$^{1}$
\qquad
Guanlin	Li$^{1}$
\qquad
Zongmin	Li$^{1}$\thanks{Corresponding author}
\\
$^{1}$ China University of Petroleum (East China) \quad $^{2}$ Polytechnic School of Chimborazo (ESPOCH)\\
{\tt\small liyachuan@s.upc.edu.cn,}
{\tt\small lizongmin@upc.edu.cn}
}
% % China\\
% {\tt\small liyachuan@s.upc.edu.cn}
% \and
% Xavier Soria Poma\\
% Polytechnic School of Chimborazo (ESPOCH)\\
% % First line of institution2 address\\
% {\tt\small xavier.soria@espoch.edu.ec}
% \and
% Yun Bai\\
% China University of Petroleum (East China)\\
% % First line of institution2 address\\
% {\tt\small baiyun@s.upc.edu.cn}
% \and
% Qian Xiao\\
% China University of Petroleum (East China)\\
% % First line of institution2 address\\
% {\tt\small xiaoqian@s.upc.edu.cn}
% \and
% Chaozhi	Yang\\
% China University of Petroleum (East China)\\
% % First line of institution2 address\\
% {\tt\small yang.chaozhi@foxmail.com}
% \and
% Guanlin	Li\\
% China University of Petroleum (East China)\\
% % First line of institution2 address\\
% {\tt\small liguanlin@s.upc.edu.cn}
% \and
% Zongmin	Li\\
% China University of Petroleum (East China)\\
% % First line of institution2 address\\
% {\tt\small lizongmin@upc.edu.cn}
% }
\maketitle

%%%%%%%%% ABSTRACT
\begin{abstract}
Transformer-based models have made significant progress in edge detection, but their high computational cost is prohibitive. 
Recently, vision Mamba have shown excellent ability in efficiently capturing long-range dependencies.
Drawing inspiration from this, we propose a novel edge detector with Mamba, termed EDMB, to efficiently generate high-quality multi-granularity edges.
In EDMB, Mamba is combined with a global-local architecture, therefore it can focus on both global information and fine-grained cues. The fine-grained cues play a crucial role in edge detection, but are usually ignored by ordinary Mamba.
We design a novel decoder to construct learnable Gaussian distributions by fusing global features and fine-grained features. And the multi-grained edges are generated by sampling from the distributions.
In order to make multi-granularity edges applicable to single-label data, we introduce Evidence Lower Bound loss to supervise the learning of the distributions.
On the multi-label dataset BSDS500, our proposed EDMB achieves competitive single-granularity ODS 0.837 and multi-granularity ODS 0.851 without multi-scale test or extra PASCAL-VOC data.  
Remarkably, EDMB can be extended to single-label datasets such as NYUDv2 and BIPED.
% The source code is available \href{https://anonymous.4open.science/r/EDMB-6F52}{here}.
The source code is available at \href{https://github.com/Li-yachuan/EDMB}{https://github.com/Li-yachuan/EDMB}.
\end{abstract}

%%%%%%%%% BODY TEXT
\section{Introduction}

Edge detection is a fundamental task in computer vision, preserving critical semantic information while mitigating texture noise. Therefore, it is widely used in downstream vision tasks such as image retrieval \cite{yu2016EdgeSketch}, salient object detection \cite{zhao2019egnet}, semantic segmentation \cite{xu2022pidnet}, and depth map prediction \cite{zhu2020edge}.

Edge detection has been studied since the last century \cite{sobel19683x3}.
Early image edge detectors \cite{davis1975survey,sharifi2002classified,sobel19683x3,canny1986computational,Hallman_Fowlkes_2014} are based on local cues, which makes it difficult for these models to understand the high-level semantic information.
With the application of deep learning in edge detection \cite{xie2017holistically,liu2019richer,he2022bdcn,soria2022ldc}, the performance of detectors has been improved unprecedented, far surpassing the local feature based methods and even surpassing human beings. Consequently,
deep learning based detectors have become the mainstream methods for current edge detection.
The alternating stacking of convolutional and pooling blocks enables the deep features to obtain larger receptive fields, which is the key to the success of deep learning-based methods. 
However, it is found that the effective receptive field is smaller than the theoretical receptive field due to the non-uniform distribution of convolution kernel weights and the inhibition effect of nonlinear activation functions \cite{luo2016understanding}. 

The self-attention mechanism, as implemented in Transformer, is introduced into edge detection to address the limitations of CNN in capturing sufficient context information \cite{pu2022edter,ye2024diffusionedge}. While self-attention effectively models global context and resolves long-range dependence, the high computational cost presents significant challenges, particularly in edge detection, which is critical for real-time applications and resource-constrained edge devices. Therefore, the principal challenge in edge detection lies in effectively capturing semantic context and modeling long-range dependence.

Another concern is multi-granularity edges.
People have different edge granularity preferences, so it is reasonable to produce different edge annotation for the same image. The preference is meaningful, and the required edge granularity is vary for different downstream tasks. For example, image segmentation \cite{xu2022pidnet} requires coarse-grained edges to refine segmentation boundary, and image retrieval \cite{yu2016EdgeSketch} requires fine-grained edges to preserve more semantic information. Therefore, it is necessary to design a multi-granularity edge detector. 
At present, the research on multi-granularity edge is in the initial stage, and the existing methods \cite{zhou2023treasure, zhou2024muge} rely on multi-label data to define the granularity of edges, which greatly limits the application scope of these methods.

To address the above issues, we propose a novel edge detector with Mamba, termed EDMB, to efficiently generate high-quality multi-granularity edges. First, we run a global Mamba encoder to capture the global context information of the image, then divide the image into non-overlapping windows and capture the fine-grained context information within the window by a fine-grained Mamba encoder. 
Although Mamba demonstrates greater efficiency compared to the Transformer, its computational cost increases proportionally with feature resolution due to the demands of long-distance modeling. With this in mind, it is advantageous to avoid applying Mamba to high-resolution features to mitigate computational cost. We propose a shared CNN-based high-resolution feature encoder to process high-resolution features, which is composed of several straightforward convolutional and nonlinear modules.
In addition, a learnable Gaussian distribution decoder is developed to fuse and decoder these features into learnable Gaussian distributions.
And the multi-grained edges are generated by sampling from the distributions. 
Evidence Lower Bound loss is introduced to supervise the distribution, which can effectively avoid the dependence on multi-label data.
%Multi-granularity edges can be obtained by sampling from the distributions.To eliminate the dependence on multi-label data and avoid variance collapse, we introduce Evidence Lower Bound loss to jointly optimize the learnable mean and variance. 

Our contributions are summarized as follows: 
\textbf{1)} We propose a novel Mamba-based edge detector, EDMB, to efficiently generate high-quality edges. To our best knowledge, EDMB is the first Mamba-based edge detector. 
\textbf{2)} We propose a multi-granularity edge generation method based on Evidence Lower Bound loss, so that the generation of multi-granularity edges does not rely on multi-label data.
\textbf{3)} Extensive experimental results show that the performance of EDMB is highly competitive on both multi-label and single-label datasets such as BSDS \cite{arbelaez2010contour}, NYUDv2 \cite{silberman2012indoor} and BIPED \cite{poma2020dense}.

\section{Related work}
\subsection{Edge Detection}

Edge detection has been studied since the last century \cite{davis1975survey,sharifi2002classified}. Early classical methods \cite{sobel19683x3, canny1986computational, Hallman_Fowlkes_2014,Dollar_Zitnick_2014} typically rely on the local cues of images.
The lack of global information limits the performance of these detectors in complex image scenes.
As the pioneer of contemporary edge detection, HED~\cite{xie2017holistically} introduces deep supervision mechanism to edge detection and learns multi-scale predictions holistically. Most of the subsequent methods \cite{liu2019richer, he2022bdcn, xuan2022fcl} follow HED and focus on feature enhancement and fusion.

In pursuit of higher accuracy, recent studies focus on two aspects. The first is to improve the representational capacity of the encoder. 
In addition to enlarge the scale of CNN \cite{zhou2023treasure,zhou2024muge},  an effective method is to introduce Transformer, which have stronger representation ability into edge detection.
EDTER \cite{pu2022edter} proposes a Transformer combining global-local features to generate high-quality edges. 
DiffusionEdge \cite{ye2024diffusionedge} treats edge detection as a denoising process and integrates Swin Transformer with diffusion to generate crisp edges.
Similarly, the success of RankED \cite{cetinkaya2024ranked} is also inseparable from the powerful representation ability of Transformer.
These methods effectively improve the performance of edge detection, but the high computational cost of Transformer limit its application in more downstream tasks.

Another research hotspot is edge uncertainty. Different people have different perception of the edge, which leads to the uncertainty of the edge and poses new challenges to edge detection.
PEdger \cite{fu2023practical} proposes a collaborative learning framework, which harnesses the disparity in edge perception between two heterogeneous models to refine the labels.
RankED \cite{cetinkaya2024ranked} introduces RS Loss to edge detection and promotes high confidence edge pixels to have more label certainty.
UAED \cite{zhou2023treasure} constructs learnable Gaussian distributions to take full use of multiple labels, and BetaNet \cite{li2023beta} replaces Bernoulli distribution with Beta distribution and employs recurrent voting strategy to merge multiple labels.
MuGE \cite{zhou2024muge} designs an edge granularity network to estimate the edge granularity from multiple labeled edge annotation, which can guide edge detector to generate edges in different granularity.
The performance of edge detection can be improved by evaluating the uncertainty of edges. In particular, UAED and MuGE can generate multi-granularity edges, which is beneficial for a wider range of applications of edge detection.
While the generation of multi-granularity edges heavily rely on multi-label data, which makes it difficult to apply these methods to the more common single-label data.

\subsection{Vision Mamba}
Mamba is a kind of selective state-space model (SSM) for modeling long-range dependencies of temporal sequences \cite{gu2023mamba}. Inspired by Vision Transformer, Mamba is applied to serialized patches of images and achieves competitive accuracy on computer vision tasks, such as image classification  \cite{zhu2024vision}, 
% object detection \cite{zubic2024state}, 
image segmentation \cite{yang2024remamber} and image restoration \cite{chen2024res}.
Mamba no longer rely on self-attention, the high computational cost module in Transformer. As a result, Mamba is significantly more efficient than Transformer, and has received continuous attention \cite{wang2024state}. 

Similar to Vision Transformer, early Vision Mamba focuses more on global modeling, which makes it less attentive to local cues. Therefore, LocalMamba \cite{huang2024localmamba} introduces a novel scanning method that consists of local scans within different windows to capture local detail information and global context. It searches for scan directions at different network layers to identify and apply the most effective combination of scans.
In addition, there are various scanning methods such as Continuous 2D Scanning, Zigzag Scan, and Multi-Path Scan \cite{zhang2024survey}, which are mainly aim to achieve the balance between the local cues and the global context of Mamba.
To date, modeling Mamba's long-range and short-range dependencies simultaneously and efficiently remains an open problem.

The computational efficiency of Mamba is between Transformer and CNN. To further improve the efficiency of the model, Mamba is combined with CNN \cite{liao2024lightm,wu2024ultralight}, where CNN is used to process high-resolution features. 
We agree with the idea and further compress the scale of the high-resolution feature encoder, which is only used to maintain the location information of features.

\section{Edge Detector with Mamba}

\begin{figure*}[!t]
\centering
\includegraphics[width=0.9\linewidth]{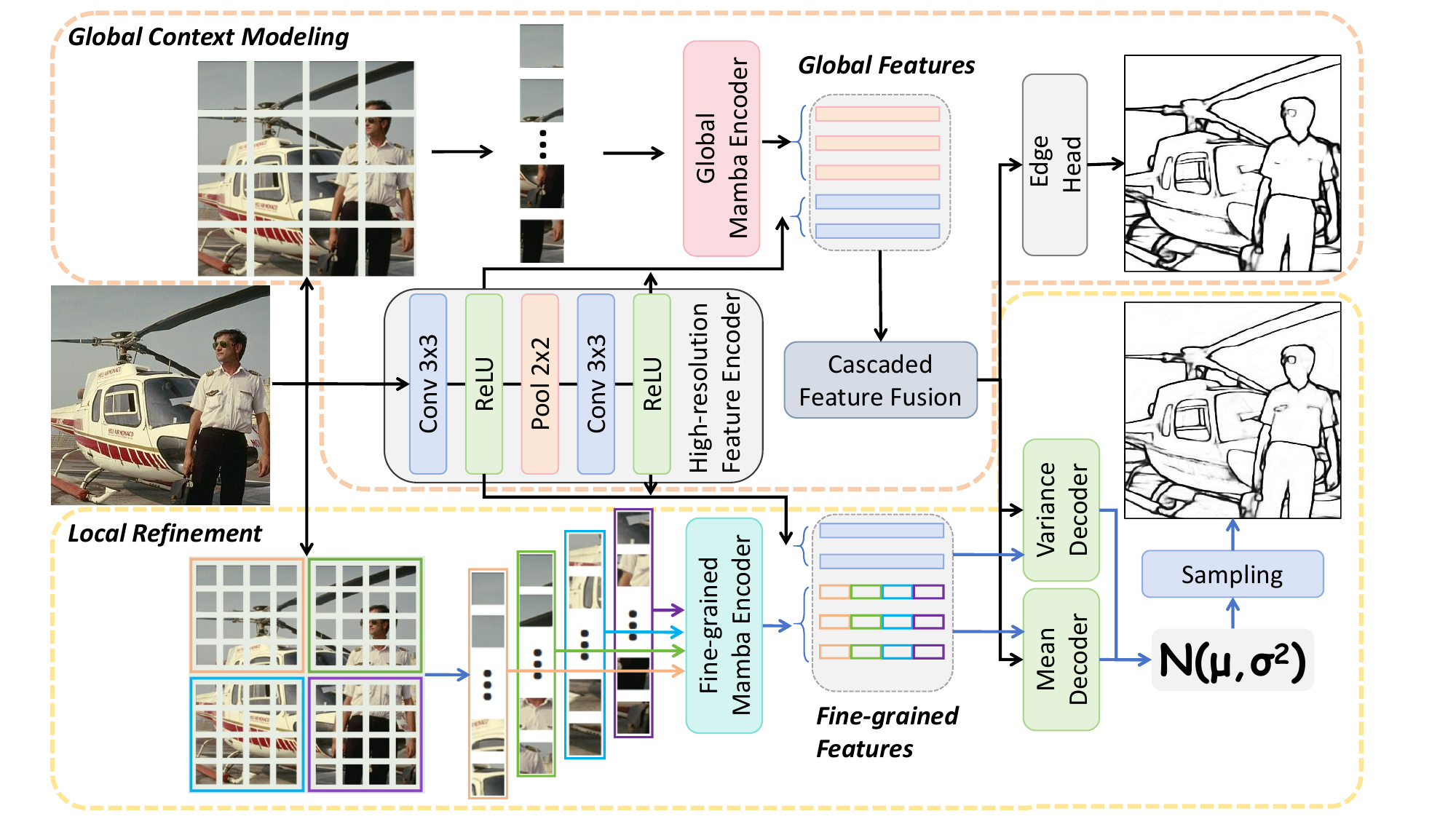}
\caption{The EDMB's framework. $N(\mu,\sigma^2)$ means learnable Gaussian distributions, where the means $\mu$ and variances $\sigma^2$ are predicted by the mean decoder and variance decoder, respectively.}
\label{fig_struct}
\end{figure*}

\subsection{Overview}

The overall framework of the proposed EDMB is illustrated in Fig. \ref{fig_struct}. 
High-resolution feature encoder $E_h$, global Mamba encoder $E_g$ and fine-grained Mamba encoder $E_f$ encode image $\mathbf{X}$ into features $\mathbf{F_h}$, $\mathbf{F_g}$ and $\mathbf{F_f}$, respectively.
Multi-scale global features $\{\mathbf{F_h}, \mathbf{F_g}\}$ and Multi-scale fine-grained features $\{\mathbf{F_h}, \mathbf{F_f}\}$ are fed into the decoder at the same time.  
Cascading feature fusion module decode $\{\mathbf{F_h}, \mathbf{F_g}\}$ into high-resolution global feature $\mathbf{F_{hg}}$, which guides the decoding of $\{\mathbf{F_h}, \mathbf{F_f}\}$.
$\{\mathbf{F_h}, \mathbf{F_f}\}$ and $\mathbf{F_{hg}}$ are combined into a new set, $\{\mathbf{F_h}, \mathbf{F_g}, \mathbf{F_{hg}}\}$, which are input into mean decoder $D_{m}$ and variance decoder $D_{v}$, respectively. 
Learnable Gaussian distributions are constructed by the output of the two decoders, and the final edge map is sampled from the distributions.

\subsection{Review Vision Mamba}

The Mamba encoders of EDMB follow the Vision Mamba (ViM) from \cite{zhu2024vision}, which is applicable in 2-D sequence. 
A 2-D image $X \in \mathbb{R}^{H \times W \times 3}$ is first transformed into flattened 2-D patches $X_p \in \mathbb{R}^{C \times (P^2 \times  3)}$, 
where $(H \times W )$ is the image size, 
$\mathtt{P}$ is the patch size. 
Next, the $X_p$ is linearly projected to latent embedding $E_l \in \mathtt{R}^{C\times N}$ and add position embeddings $\mathbf{E}_{pos} \in \mathbb{R}^{C \times \mathtt{N}}$.
Different from ViT \cite{dosovitskiy2020image}, which directly computes the correlation between all patches through costly Self-Attention, the core of ViM is advanced state space model (SSM).
In SSM, the correlation between patches is established through the state matrix $\mathbf{A}\in \mathbb{R}^{N\times N}$ and projection parameters $\mathbf{B}\in \mathbb{R}^{C\times 1}$, $\mathbf{C}\in \mathbb{R}^{C\times 1}$.
The process can be represented as
\begin{equation}
\begin{aligned}
\label{eq:lti}
h'(t) &= \mathbf{A}h(t) + \mathbf{B}x(t), \\
y(t) &= \mathbf{C}h(t)
\end{aligned}
\end{equation}
where $h(t) \in \mathbb{R}^N$ represents intermediate implicit states.
To transform the continuous parameters $\mathbf{A}$, $\mathbf{B}$ to discrete parameters $\mathbf{\overline{A}}$, $\mathbf{\overline{B}}$, transformation method zero-order hold is used:
\begin{equation}
\begin{aligned}
\label{eq:zoh}
\mathbf{\overline{A}} &= \exp{(\mathbf{\Delta}\mathbf{A})}, \\
\mathbf{\overline{B}} &= (\mathbf{\Delta} \mathbf{A})^{-1}(\exp{(\mathbf{\Delta} \mathbf{A})} - \mathbf{I}) \cdot \mathbf{\Delta} \mathbf{B}.
\end{aligned}
\end{equation}

% And Eq. \ref{eq:lti} with step size $\mathbf{\Delta}$ can be rewritten as:
% \begin{equation}
% \begin{aligned}
% \label{eq:discrete_lti}
% h_t &= \mathbf{\overline{A}}h_{t-1} + \mathbf{\overline{B}}x_{t}, \\
% y_t &= \mathbf{C}h_t.
% \end{aligned}
% \end{equation}

The output $\mathbf{y}$ of SSM is compute through 
$\mathbf{y} = \mathbf{x} *  (\mathbf{C}\mathbf{\overline{B}}, \mathbf{C}\mathbf{\overline{A}}\mathbf{\overline{B}}, \dots, \mathbf{C}\mathbf{\overline{A}}^{M-1}\mathbf{\overline{B}})$,
where $M$ is the length of the input sequence $\mathbf{x}$.
Vanilla SSM is unidirectional correlation, which is suitable for temporal sequence. ViM constructs bidirectional SSM, $SSM_{f}(\cdot)$ and $SSM_{b}(\cdot)$, by reversing the input sequence to capture the context information between image patches.
The final result can be roughly expressed as:
\begin{equation}
\mathbf{T_l} = Lin(SSM_{f}(\mathbf{T_{l-1}})+SSM_{b}(\mathbf{T_{l-1}}))+\mathbf{T_{l-1}},
\end{equation}
where $T$ is the token sequence. $l$ is the l-th layer of the Vim encoder. $Lin(\cdot)$ is a linear projection.

\subsection{Mamba-based Encoder}
% The success of ViM is undeniable, achieving comparable accuracy to ViT with less computational cost, but it has the following problems when applied to edge detection:
% \textbf{1)} While ViM is less computationally expensive than ViT, it is significantly more expensive than CNN, especially for high-resolution features.
% \textbf{2)} Modeling fine-grained features is important for edge detection, which is ignored by ViM.
% \textbf{3)} Low-resolution features contribute little to the accuracy, but the computational cost is not negligible.

% To solve the above problems, we propose a novel Mamba-based encoder as shown in Fig. \ref{fig_struct}. 

The Mamba-based encoder is shown in Fig. \ref{fig_struct}, where features are divided into global features, fine-grained features, and high-resolution features. 
Those features are encoded by global Mamba encoder $E_g$ , fine-grained Mamba encoder $E_f$ and high-resolution feature encoder $E_h$, respectively.
Following ViM, $E_g$ performs 4$\times$ downsampling during image patching to avoid Mamba's high computational cost on high-resolution features. And only the first three stages of the ViM encoder are adopted, avoiding meaningless computations on low-resolution features.
Specifically, the input image is split into a sequence of $4\times 4$ patches, and then the sequence is linearly projected to latent embedding.
The output $\mathbf{F_g}$ of $E_g$ can be compute by
\begin{equation}
    \mathbf{F_g} = \{\mathbf{F_g^1},\mathbf{F_g^2},\mathbf{F_g^3}\}= E_g(\mathbf{X}),
\end{equation}
where $\mathbf{F_g^i}$ is the output of stage $i$ in $E_g$.

% Fine-grained context features is essential for edge detection, while it is ignored by $E_g$.
Inspired by EDTER \cite{pu2022edter}, we propose fine-grained Mamba encoder $E_f$ to capture fine-grained cues which is ignored by $E_g$. 
Although $E_f$ has the same structure as $E_g$, the data processing is different.
%
% In particular, 
We decompose image $X \in \mathbb{R}^{H \times W \times 3}$ into a sequence $\{X_1, X_2, X_3, X_4\}$ through a $\frac{H}{2} \times \frac{W}{2}$ non-overlapping sliding window and encode the sequence with the fine-grained Mamba encoder.
We randomly drop the gradient of $X_i$ to reduce the computational cost during training. 
While the whole sequence is encoded to obtain the full fine-grained image features $F_f$. 
The output $\mathbf{F_f}$ of $E_f$ can be compute by
\begin{equation}
    \mathbf{F_f} = \{\mathbf{F_f^1},\mathbf{F_f^2},\mathbf{F_f^3}\}= E_f(\mathbf{\{X_1, X_2, X_3, X_4\}}).
\end{equation}

Compared to CNN, Mamba is more expensive to compute on high-resolution features, therefore, the data is downsampled before input to Mamba-based encoders. In order to make up for the location error caused by downsampling with a smaller cost, we propose high-resolution feature encoder $E_h$, which is a lightweight encoder based on CNN.
$E_h$ contains two vanilla Convolution blocks, and each vanilla convolution block consists of a $3\times 3$ convolution layer and a ReLU activation. 
To ensure the accuracy of the location information, the feature resolution in the first vanilla Convolution block is consistent with the input image.
A down-sampling module is inserted between the two convolution blocks to generate two-scale location features. 
Consistent with VGG16~\cite{simonyan2014very}, the number of channels doubles when the feature resolution is halved.
Since accurate location information rather than rich semantic information is mainly consider, the features in $E_h$ is compressed to 16 channels, which is much smaller than existing models such as VGG16~\cite{simonyan2014very} (64 channels) and ResNet50~\cite{he2016deep} (64 channels). 
Compared to VGG16 and ResNet50, $E_h$ not only has fewer modules, but also has fewer channels.
Therefore, even though the features keep high resolution, the computational cost remains low. The output $\mathbf{F_h}$ of $E_h$ can be computed by 
\begin{equation}
    \mathbf{F_h} = \{\mathbf{F_h^1},\mathbf{F_h^2}\}= E_h(\mathbf{X}).
\end{equation}

\subsection{Learnable Gaussian Distributions Decoder}
%损失函数通常被称为“变分下界”（Variational Lower Bound），也常被称为 ELBO（Evidence Lower BOund）。ELBO 

\begin{figure*}[!t]
\centering
\includegraphics[width=0.9\linewidth]{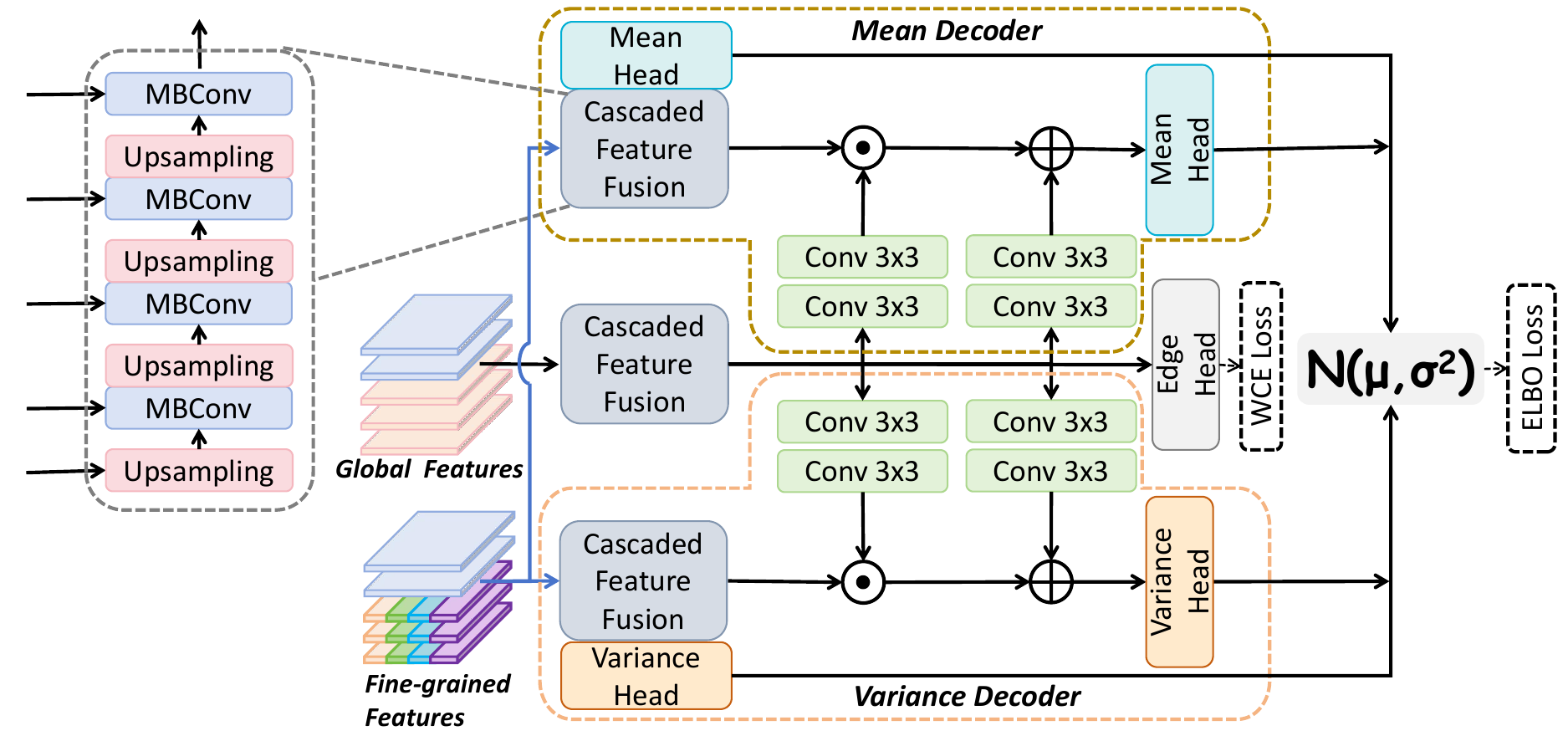}
\caption{The detailed architecture of the LGD decoder. $MBConv$ is a kind of efficient convolution proposed by MobileNetv2 \cite{howard2017mobilenets}.}
\label{fig_decoder}
\end{figure*}

Following previous works \cite{zhou2023treasure,zhou2024muge}, the key to generate multi-granularity edges is to construct learnable Gaussian distributions, and adjust the edge granularity by changing the variance.
Therefore, We propose a learnable Gaussian distributions (LGD) decoder that utilizes multi-scale features to predict the means and variances of learnable Gaussian distributions, as shown in fig. \ref{fig_decoder}.

The input of the decoder is multi-scale global features $\{\mathbf{F_g},\mathbf{F_h}\}$ and multi-scale fine-grained features $\{\mathbf{F_f},\mathbf{F_h}\}$.
A Cascaded Feature Fusion (CFF) module fuses $\{\mathbf{F_g},\mathbf{F_h}\}$ to obtain high-resolution global features $\mathbf{F_{hg}}$. 
The other two CFF modules utilize $\{\mathbf{F_f},\mathbf{F_h}\}$ to generate fine-grained mean features $\mathbf{F_{fm}}$ and fine-grained variance features $\mathbf{F_{fv}}$, respectively.
The structure of CFF module is shown in the left part of fig. \ref{fig_decoder}. 
The upsampled low-resolution features are concatenated with the adjacent higher-resolution features in the channel dimension, the result are fused by MBConv \cite{howard2017mobilenets}. The output of the module is the features with highest resolution.
CFF module is widely applied \cite{ronneberger2015u} due to its balance in effectiveness and efficiency.
Global features contain richer global context for distinguishing edges and textures more effectively from the semantic level. Fine-grained features are better at capturing short-range local cues and are able to perceive thinner edges. Therefore, $\mathbf{F_{hg}}$ is combined with $\mathbf{F_{fm}}$ and $\mathbf{F_{fv}}$ to generate the mean and variance respectively in the mean decoder $D_m$ and variance decoder $D_v$. 

Following EDTER \cite{pu2022edter}, a spatial feature transform block is hired in $D_m$ to fully modulate the global features and fine-grained features. 
A mean head consisting of several simple Conv-BN-ReLU modules maps the results to single-channel means of distributions.
$D_v$ is basically the same as $D_m$ in structure. The difference is that the variance head contains an additional SoftPlus function to process the output to ensure the non-negativity of the variances of distributions.

In addition, we hire an Edge Head, a Mean Head and a Variance Head to directly deal with high-resolution global features $\mathbf{F_{hg}}$, fine-grained mean features $\mathbf{F_{fm}}$ and fine-grained variance features $\mathbf{F_{fv}}$,  which are used to aid training and will be discarded during inference. The outputs of the three auxiliary head are termed $p_a$, $\mu_a$ and $\sigma_a^2$.
The learnable Gaussian distributions can be expressed as
\begin{equation}
\begin{aligned}
    \mathcal{N}(\mu,\sigma^2) 
    &= \mathcal{N}(D_{LGD}(\{\mathbf{F_g},\mathbf{F_h}\},\{\mathbf{F_f},\mathbf{F_h}\}))\\
    &= \mathcal{N}(D_m(\mathbf{F_{hg}},\mathbf{F_{fm}}),D_v(\mathbf{F_{hg}},\mathbf{F_{fv}}))
\end{aligned}
\end{equation}
% where $D_m$ means Mean Decoder and $D_v$ means Variance Decoder.

\subsection{Loss Function}

To force the outputs obey Gaussian distributions, we introduce ELBO (Evidence Lower BOund) loss \cite{kingma2013vae} to supervise the learning of EDMB. 
ELBO loss consists a Kullback-Leibler (KL) divergence, which measures the similarity between the predicted distribution and the prior distribution, and a Weighted Cross Entropy (WCE) loss, which measures the distance between the ground truth $y$ and a sampling $p$ of the distributions.

For image $X$, KL loss can be formulated as
\begin{equation}
\begin{aligned}
L_{kl}&(N(\mu, \sigma^2),N(0,1)) \\
= & \int \frac{1}{\sqrt{2 \pi \sigma_i^2}} e^{\xi}\left(\log \frac{e^{\xi} / \sqrt{2 \pi \sigma_i^2}}{e^{-x^2 / 2} / \sqrt{2 \pi}}\right) dx \\
= & \int \frac{1}{\sqrt{2 \pi \sigma_i^2}} e^{\xi}\log \left\{\frac{1} 
{\sqrt{\sigma_i^2}} \exp \left\{\frac{1}{2}\left(x^2+ 2 \xi \right)\right\}\right\} d x \\
= & \frac{1}{2} \int \frac{1}{\sqrt{2 \pi \sigma_i^2}} e^{\xi}\left(-\log \sigma_i^2+x^2 + 2 \xi \right)dx\\
= &\frac{1}{2} \sum_{i=1}^{N}(\mu_i^2+\sigma_i^2-\log \sigma_i^2-1),
\end{aligned}
\end{equation}
where \(\xi = -(x-\mu_i)^2 / 2 \sigma_i^2\) and $N$ is the total number of pixels in image $X$.
WCE loss \cite{zhou2023treasure,zhou2024muge} can be formulated as
\begin{equation}
    L_{wce}(p,y)=-\sum_{i=1}^{N} \alpha \log \left(p_{i}\right) + \beta \log \left(1-p_{i}\right),
    \label{eq-wce}
\end{equation}
where \textit{p} denotes a sampling $p$ of the learnable distributions $N(\mu, \sigma^2)$ and \textit{y} is the ground truth. $\alpha=\lvert Y_{+} \rvert/\lvert Y \rvert$ and $\beta =\lambda\cdot\lvert Y_{-} \rvert/\lvert Y \rvert$. $\lvert Y_{+} \rvert$ and $\lvert Y_{-} \rvert$ are used to represents the number of edge and non-edge, separately. $\lambda$ controls the weight of positive over negative samples. $\lvert Y \rvert=\lvert Y_{+} \rvert + \lvert Y_{-} \rvert$. $N$ is the total number of pixels in image $X$.
Therefore, ELBO loss is defined as
\begin{equation}
\begin{aligned}
L_{elbo}&(p,y,N(\mu, \sigma^2),N(0,1))\\
=&L_{wce}(p,y) + \varphi L_{kl}(N(\mu, \sigma^2),N(0,1)),
\end{aligned}
\end{equation}
where $\varphi$ is a hyperparameter to balance the reliability and randomness of the output.

In practice, we easily encounter the predicament that global features and fine-grained features do not converge at the start of training. 
We propose two auxiliary losses to supervise the learning of global features and fine-grained features.
As shown in Fig \ref{fig_decoder}, WCE loss $L_{wce}(p_a,y)$ is used for the high-resolution global features $\mathbf{F_{hg}}$. 
ELBO loss $L_{elbo}(p_a,y,N(\mu_a, \sigma_a^2),N(0,1))$ is used for the fine-grained mean features $\mathbf{F_{fm}}$ and the fine-grained variance features $\mathbf{F_{fv}}$.

The total loss of EDMB is 
\begin{equation}
\begin{aligned}
L=&L_{elbo}(p,y,N(\mu, \sigma^2),N(0,1)) \\
  &+ \alpha_1 L_{wce}(p_a,y) \\
  &+\alpha_2 L_{elbo}(p_a,y,N(\mu_a, \sigma_a^2),N(0,1)),
\end{aligned}
\label{eq-loss}
\end{equation}
where $\alpha_1$ and $\alpha_2$ are hyperparameters to balance multiple loss functions.

\subsection{Training Strategy}

In theory, EDMB can be trained end-to-end. While we follow the practice of EDTER \cite{pu2022edter} and adopt multi-stage training. 
As shown in Fig \ref{fig_struct}, the global Mamba encoder is employed at first to catch the global context, and the loss function in this stage is
\begin{equation}
    L_g= L_{wce}(p_a,y).
    \label{loss-stage1}
\end{equation}

Then, the global information can guide the learning of the fine-grained Mamba encoder and generate learnable Gaussian distributions (LGD). The loss function in this stage is
\begin{equation}
\begin{aligned}
L_f=&L_{elbo}(p,y,N(\mu, \sigma^2),N(0,1)) \\
  &+\alpha_2 L_{elbo}(p_a,y,N(\mu_a, \sigma_a^2),N(0,1)).
\end{aligned}
\label{loss-stage2}
\end{equation}

Compared with end-to-end training, multi-stage training has three main advantages: 
\textbf{1)} In the local refinement stage, gradients are not applied to the global Mamba encoder or the high-resolution feature encoder, in order to minimize computational overhead and storage demands.
\textbf{2)} The gradient of the model is more stable and makes model training easier. 
\textbf{3)} Compared to Eq. \ref{eq-loss}, optimizing Eq. \ref{loss-stage1} and Eq. \ref{loss-stage2}, respectively, avoids the dependence on the hyperparameter $\alpha_1$.

Different edges can be obtained by sampling independently multiple times from the learnable Gaussian distributions, but they are in the same granularity. To generate multi-granularity edges, we follow the practice of MuGE \cite{zhou2024muge} and control the granularity of the edges by adjusting the weight of the learned variance. The final edge map can be formulated as
\begin{equation}
    \label{eq-sampling}
    p_\gamma = \mu+\gamma\sigma^2, 
\end{equation}
where $\gamma$ is a hyperparameter that controls the granularity.

\section{Experiments}

\subsection{Datasets} 
\label{sec-dataset}
We evaluate the proposed EDMB on three widely used datasets BSDS500, NYUDv2, and BIPED.
\textbf{BSDS500}~\cite{arbelaez2010contour} is a popular natural scenes dataset for edge detection,  consisting of 200 training images, 100 validation images, and 200 test images. Each image is annotated by 4 to 9 experts. In accordance with previous studies~\cite{xie2017holistically,liu2019richer}, the model is trained on both the training and validation sets, and evaluated on the test set.
\textbf{NYUDv2}~\cite{silberman2012indoor} is a indoor scene semantic segmentation dataset. The edge ground truth is generated from the segmentation maps. NYUDv2 comprises 1,449 groups of meticulously annotated RGB and depth images, with each group containing a single annotated edge. For our experiments, we utilized 795 images for training the model and evaluated its performance on the remaining images.
\textbf{BIPED}~\cite{soria2023dexinedv2} contains 250 carefully annotated high-resolution Barcelona Street View images. There are 200 images for training, and the rest 50 images for testing. All images are carefully annotated at single-pixel width by experts in the computer vision. The data augmentation strategy during training is shown in Table~\ref{tab:dataAug}. 
% The data augmentation strategy on BSDS500, NYUDv2 and  BIPED
% of BSDS is consistent with LPCB~\cite{deng2018learning}. The data augmentation strategy of NYUDv2 follows RCF~\cite{liu2019richer}, with the only difference that the augmentation is randomly cropped to $400\times400$ for batch processing. The data augmentation strategy of BIPED is consistent with DexiNed~\cite{poma2020dense}.
% The data augmentation on BSDS500, NYUDv2, and BIPED follow LPCB \cite{deng2018learning}, RCF \cite{liu2019richer}, and DexiNed \cite{poma2020dense}, respectively. And additional $400\times400$ clipping is applied to NYUDv2 for batch processing.

\begin{table}[ht]
\setlength{\tabcolsep}{2pt} % 减小列间隔
    \centering
    \begin{tabular}{lcc}
        \hline
        Datasets & Augmentation strategies& Resize\\
        \hline
         BSDS& F ($4\times$), R ($25\times$) & $481\times321$\\
         NYUDv2& F ($2\times$), S ($3\times$), R ($4\times$)&$400\times400$\\
         BIPED&F ($2\times$), S ($3\times$), R ($16\times$), G ($3\times$)&$400\times400$\\
         \hline
    \end{tabular}
    \caption{Augmentation strategies adopted on three edge detection benchmarks. \textbf{\textit{F}}: flipping, \textbf{\textit{S}}: scaling, \textbf{\textit{R}}: rotation, \textbf{\textit{C}}: cropping, \textbf{\textit{G}}: gamma correction.}
    \label{tab:dataAug}
\end{table}

\subsection{Implementation Details}

The proposed EDMB is implemented using PyTorch library. All parameters are updated by Adam optimizer. The Mamba based encoders are pre-trained on ImageNet, and the remaining modules are randomly initialized. The batchsize is 4 for the global context modeling stage. And the batchsize of the local refinement stage is reduced to 3 due to the limitation of the memory.
Weight decay is set to 5e-4. For the pretrained module, the learning rate is 1e-5, and the rest is set to 1e-4. 
Following previous works \cite{liu2019richer,poma2020dense}, the hyperparameter $\lambda$ is set to 1.1 for both BSDS500 and BIPED, and 1.3 for NYUDv2.
The hyperparameter $\alpha_2$ in Eq. \ref{loss-stage2} is set to 0.4 for balancing the multiple loss functions.
A standard Non-Maximum Suppression (NMS) is performed to produce the final edge maps before the quantitative evaluation. $F_{-} measure=\frac{2\times P\times R}{P+R}$ is utilized as the quality evaluation standard of the generated edge map, where \emph{P} represents the accuracy rate and \emph{R} represents the recall rate. 
The edges are binarized by an optimal threshold value for each image (OIS) or the same optimal threshold for the whole dataset (ODS). 
For more experimental details, please refer to previous works \cite{liu2019richer,poma2020dense}.

In addition to accuracy measurements, key metrics are included to evaluate the model compactness and inference efficiency. They are measured by model size (Param) and Giga Floating-Point Operations (GFLOPs).
GFLOPs is a hardware-independent metric of computational requirements that is more fair compared to Frames Per Second (FPS).

\subsection{Comparison with State-of-the-arts}

\begin{table}[ht]
\setlength{\tabcolsep}{2pt} % 减小列间隔
    \centering
    \begin{tabular}{lccccc}
    \hline
         Methods& ODS& OIS & Param&  GFLOPs\\
         \hline
         Canny \textsubscript{TPAMI'86}~\cite{canny1986computational}& 0.611     & 0.676 &-&- \\
         gPb-UCM \textsubscript{TPAMI'10}~\cite{arbelaez2010contour}  & 0.729     & 0.755  &-&- \\
         SCG \textsubscript{NeurIPS'12}~\cite{ren2012discriminatively} & 0.739     & 0.758   &-&-\\
         SE \textsubscript{TPAMI'14}~\cite{dollar2014fast} & 0.743     & 0.764 &-& -\\
         OEF \textsubscript{CVPR'15}~\cite{hallman2015oriented}      & 0.746     & 0.770   &-&- \\
         \hline
         % DeepEdge \textsubscript{CVPR'15}~\cite{bertasius2015deepedge}   & 0.753     & 0.772   & - &- \\
         DeepContour \textsubscript{CVPR'15}~\cite{shen2015deepcontour} & 0.757     & 0.776 &-&-\\
         Deep Boundary \textsubscript{ICLR'15}~\cite{kokkinos2015pushing} &0.789&0.811   &-&-\\
         CEDN \textsubscript{CVPR'16}~\cite{yang2016object}      & 0.788     & 0.804  &- &- \\
         RDS \textsubscript{CVPR'16}~\cite{liu2016learning}      & 0.792     & 0.810  &  - &- \\
         AMH-Net \textsubscript{NeurIPS'17}~\cite{xu2017learning}  & 0.798     & 0.829     &-&-\\
         HED \textsubscript{TPAMI'17}~\cite{xie2017holistically}      & 0.788     & 0.808   &14.6&57.5\\
         CED \textsubscript{CVPR'17}~\cite{wang2017deep}   &0.803&0.820  &21.8&-\\
         LPCB \textsubscript{ECCV'18}~\cite{deng2018learning}    &0.800&0.816  &-&- \\
         RCF \textsubscript{TPAMI'19}~\cite{liu2019richer}   &0.798&0.815  &14.8&75.3 \\
         DSDC \textsubscript{ ACMMM'20}~\cite{deng2020deep} &0.802&0.817  &-&- \\
         LDC \textsubscript{ACMMM'21}~\cite{deng2021learning}   &0.799&0.816  &-&-\\
         BDCN \textsubscript{TPAMI'22}~\cite{he2022bdcn}   &0.806&0.826 &16.3&103.4\\
         FCL \textsubscript{NN'22}~\cite{xuan2022fcl}   &0.807&0.822  &16.5&134.4\\
         EDTER \textsubscript{CVPR'22}~\cite{pu2022edter}  & 0.824 &0.841  &468.8&802.3 \\
         PiDiNet \textsubscript{TPAMI'23}~\cite{su2023lightweight}      & 0.789 & 0.803&0.71&11.6\\
         UAED \textsubscript{ CVPR'23}~\cite{zhou2023treasure}& 0.829  & 0.847 &72.5&72.8\\
         RankED \textsubscript{CVPR'24}~\cite{cetinkaya2024ranked} &0.824 &0.840 &121.0 & 181.0\\
         % MuGE~\cite{zhou2024muge}&CVPR'24 &0.831&0.847&113/72.5&120/73\\
         MuGE \textsubscript{CVPR'24}~\cite{zhou2024muge} &0.831&0.847&72.5&73\\
         DiffusionEdge \textsubscript{AAAI'24}~\cite{ye2024diffusionedge} & \second{0.834} & \second{0.848} &225&494\\
         EDMB (Our) & \best{0.837} & \best{0.851} &71.0&266.3\\
         \hline
         UAED*&   0.841  & 0.847 &72.5&-\\
         MuGE*&\second{0.850}&\best{0.856}&72.5&-\\ 
         EDMB* (Our)  & \best{0.851} & \second{0.854} &71.0&-\\
    \hline
    \end{tabular}
    \caption{Quantitative results on the BSDS500 dataset. All edges are generated without multi-scale testing or additional PASCAL-VOC data. * means generating multi-granularity edges. 
    The best two results are denoted as \best{red} and \second{blue} respectively, and the same for other tables.}
    \label{tab:rsult-bsds}
\end{table}

We compare our method with classic edge detectors (Canny~\cite{canny1986computational}, SCG~\cite{ren2012discriminatively}, OEF~\cite{hallman2015oriented}, \etc), CNN-based detectors (RCF~\cite{liu2019richer}, BDCN~\cite{he2022bdcn}, UAED~\cite{zhou2023treasure}, \etc) and the Transformer-based detectors (EDTER~\cite{pu2022edter}, DiffusionEdge \cite{ye2024diffusionedge}, \etc) on BSDS500. 
Quantitative results are shown in Table~\ref{tab:rsult-bsds}. By training on the train-val set of BSDS500, our method achieves the F-measure ODS of 0.837 with single-granularity testing, which is superior to all the state-of-the-art edge detectors. 
Compared with the second best method DiffusionEdge, both ODS and OIS of EDMB are improved by 0.3\%. More importantly, the model size of EDMB is only 1/3 of that of DiffusionEdge, and the computational cost is reduced to 1/2 of that of DiffusionEdge. It reflects the advantages of Mamba-based models over Transformer-based methods in edge detection.

Following MuGE \cite{zhou2024muge}, we evaluate the multi-granularity edge of EDMB and compare it with UAED and MuGE, which is denoted as EDMB*, UAED* and MuGE*.
By tuning the hyperparameter $\gamma$ in Eq. \ref{eq-sampling}, we can obtain multi-granularity edges. EDMB* generates M = 11 samples with $\gamma = \frac{n}{2}-5,n\in\{0\sim 10\}$. 
The sampling number M of EDMB* is the same as MuGE* and UAED* for fair comparison.
EDMB* achieves the best ODS of 0.851, surpassing UAED* and MuGE*.
Although compared with MuGE*, EDMB* has no significant advantage in accuracy but higher computational cost. However, EDMB* gets rid of the dependence on multi-label data, which makes EDMB* can be applied to a wider range of single-label data. Moreover, the computational cost of multi-granularity edges is proportional to the granularity M, which is much higher than that of single-granularity edges. Therefore, the computational cost is usually not considered when evaluating multi-granularity edges.

The experimental results on the single-label datasets NYUDv2 and BIPED are recorded in Table \ref{tab:result-others}.
The ODS of EDMB* on NYUDv2 is 0.783, which can match the SOTA method RankED and significantly better than other methods.
On BIPED, EDMB* achieves 0.924 ODS, which is the new SOTA method and 0.7\% higher than the second best DiffusionEdge. The improvement in accuracy is astounding.

\begin{figure*}[!t]
\centering
\includegraphics[width=\linewidth]{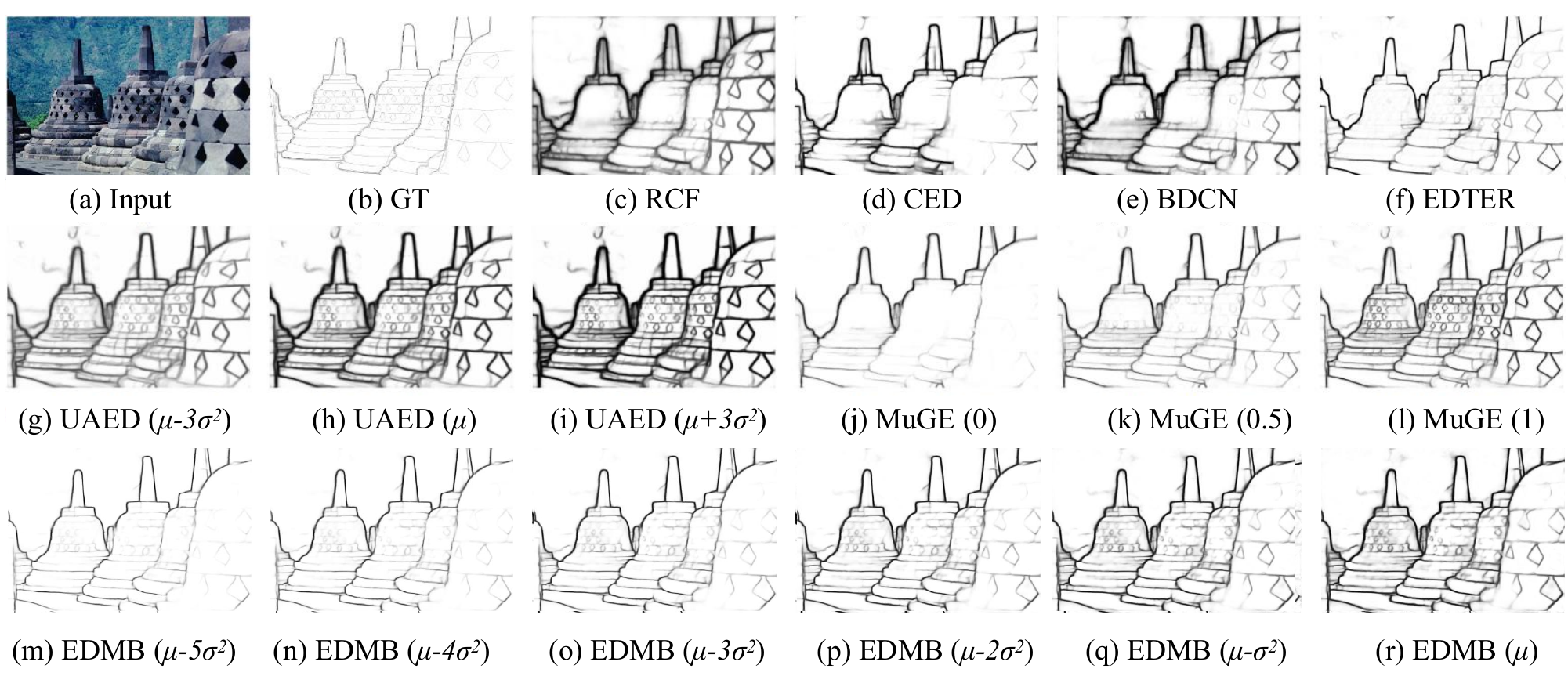}
\caption{Qualitative comparisons on challenging samples in the BSDS500 test set. MuGE produces diverse results with edge granularity of 0, 0.5, and 1, respectively.
EDMB and UAED samples from the learned distribution with $\mu+\gamma\sigma^2$. Fig. (c)$ \sim $(l) come from \cite{zhou2024muge}.}
\label{fig:bsds-vis}
\end{figure*}

Qualitative results are presented in Fig.~\ref{fig:bsds-vis}. 
Overall, EDMB* gets crisper edges. 
The granularity of EDMB* is similar to that of MuGE*, and is significantly better than that of UAED*.

%
% Some qualitative results are shown in Fig.~\ref{fig:other-vis}. 
% We can observe that on these two datasets, the granularity variation range of edges is significantly smaller than the results of BSDS500. This is mainly caused by the dataset characteristics, the images of these two datasets mainly focus on artificial buildings, and these objects have obvious edges and low uncertainty, so the multi-granularity results of EDMB are not obvious. Nonetheless, we are still able to observe a distinction between edges of different granularities.

\begin{table}
\setlength{\tabcolsep}{5pt} % 减小列间隔
    \centering
    \begin{tabular}{lcccc}
    \hline
    \multirow{2}*{Methods}& \multicolumn{2}{c}{NYUDv2} & \multicolumn{2}{c}{BIPED} \\
         \cline{2-5}
         & ODS & OIS &ODS & OIS \\
         \hline
         AMH-Net \textsubscript{NeurIPS'17} & 0.744& 0.758 &-&-  \\
         HED \textsubscript{TPAMI'17}  &  0.722 & 0.737& 0.829 & 0.847   \\
         RCF \textsubscript{TPAMI'19}  &  0.745& 0.759&  0.843 &0.859    \\
         BDCN \textsubscript{TPAMI'22}  &  0.748 &0.762&  0.839 &0.854   \\
         PiDiNet \textsubscript{ICCV'21}  & 0.733 & 0.747 & 0.868& 0.876   \\
         EDTER \textsubscript{CVPR'22}   &  0.774 & 0.789&  0.893& 0.898  \\
         DexiNed \textsubscript{PR’23} &-      &-     &0.895  &0.900    \\
         DiffusionEdge \textsubscript{AAAI'24}   &  0.761 & 0.766& 0.899& 0.901  \\
         RankED \textsubscript{CVPR'24}&\second{0.780}&\best{0.793}&-&-\\
         EDMB (Our)  & 0.774&0.787 & \second{0.903} & \second{0.907} \\
         EDMB* (Our)  & \best{0.783}&\second{0.789} & \best{0.906} & \best{0.908} \\
    \hline
    \end{tabular}
    \caption{Quantitative comparisons on NYUDv2 and BIPED. All results are computed with a single scale input. * means generating multi-granularity edges.}
    \label{tab:result-others}
    
\end{table}

% \begin{figure*}[!t]
% \centering
% \includegraphics[width=\linewidth]{z_othdset_vis.pdf}
% \caption{Multi-granularity edges of two challenging samples in NYUDv2 and BIPED test sets.}
% \label{fig:other-vis}
% \end{figure*}

\subsection{Ablation Study}

\begin{table}[!t]
\centering
\begin{tabular}{ccccccc}
\hline
$E_g$      &   $E_f$    &  $E_h$   &    LGD    &   MG              &ODS    &   OIS\\
\hline
\checkmark  &            &          &           &                   &0.821  &   0.830\\
\checkmark  &\checkmark  &          &           &                   &0.836  &   0.846\\
% 3&            &\checkmark  &\checkmark&           &                   &0.823  &   0.829\\
\checkmark  &\checkmark  &\checkmark&           &                   &\second{0.839}  & 0.848\\
\checkmark  &\checkmark  &\checkmark&\checkmark &                   &0.837  & \second{0.851}\\
% 6&            &\checkmark  &\checkmark&\checkmark &\checkmark         &0.828  &   0.837\\
\checkmark  &\checkmark &\checkmark &\checkmark &\checkmark         &\best{0.851}& \best{0.854}\\
\hline
\end{tabular}
\caption{Ablation study of each component on BSDS500.}
\label{tab-abl}
\end{table}

We perform ablation experiments on BSDS500, and the results are recorded in Table \ref{tab-abl}.
$E_g$, $E_f$ and $E_h$ mean the global Mamba encoder, the fine-grained Mamba encoder and the high-resolution feature encoder.
The contribution of each module can be visualized by the improvement of accuracy. Although LGD does not directly contribute to the accuracy, the construction of LGD is a prerequisite for multi-granularity (MG) edge and cannot be omitted.

% $\mu+i\sigma^2$, where i = 0,  $\pm1.5$, $\pm3$, $\pm4.5$, $\pm6$ and $\pm7.5$

\begin{table}[!t]
\setlength{\tabcolsep}{4pt} % 减小列间隔
\centering
\begin{tabular}{ccccc}
\hline
Method&Label      &   Granularity ($\gamma$)    &ODS    &   OIS\\
\hline
UAED    & Random&$\pm \frac{n}{2}, n\in\{0,2\sim6\}$        &0.841&0.847\\
EDMB    & Random& $\pm \frac{n}{2}, n\in\{0,2\sim6\}$       &\second{0.849}&\best{0.855}\\
EDMB    & Mixed&  $\pm \frac{n}{2}, n\in\{0,2\sim6\}$       &0.847&0.853\\
EDMB    & Mixed& $\frac{n}{2}-5,n\in\{0\sim 10\}$           &0.845&0.852\\
EDMB    & Random& $\frac{n}{2}-5,n\in\{0\sim 10\}$          &\best{0.851}&\second{0.854}\\
\hline
\end{tabular}
\caption{Effects of multi-label handling methods and granularity choices on BSDS500 dataset. Multi-granularity is obtained by sampling $\mu+\gamma\sigma^2$ from LGD.}
\label{tab:abl-mlmg}
\end{table}

\begin{figure}[!t]
\centering
\includegraphics[width=\linewidth]{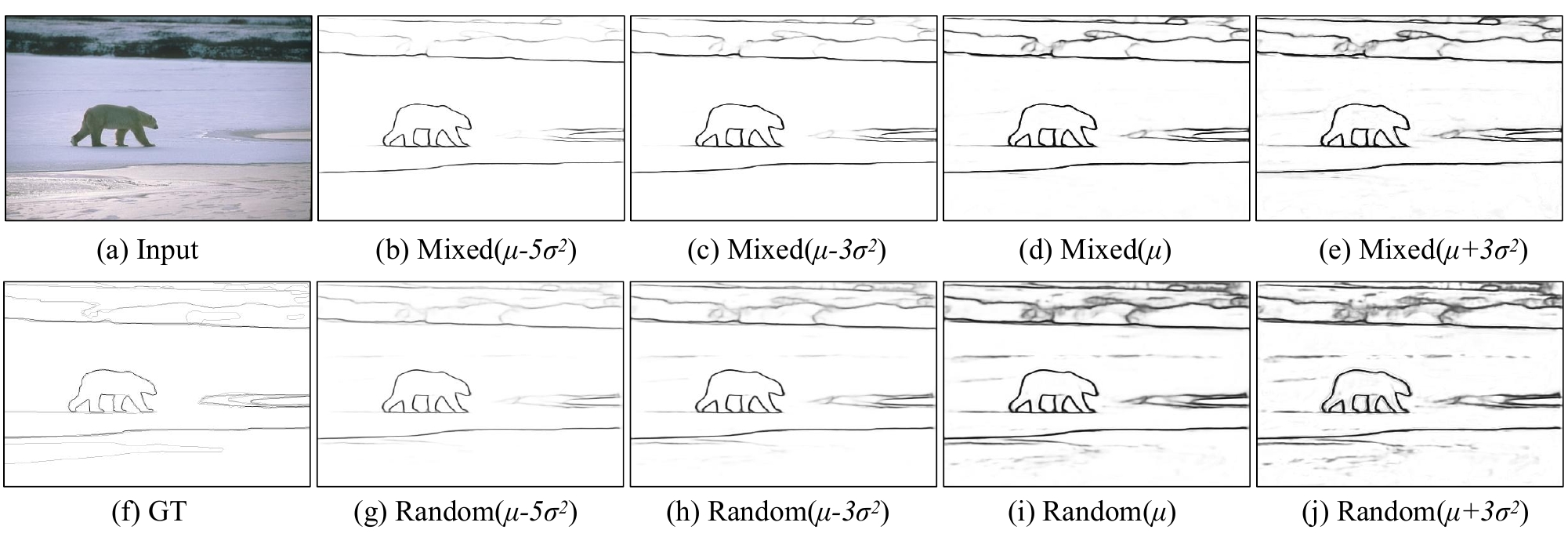}
\caption{Results of different multi-label handling methods. The first row shows the results for the mixed label, and the second row shows the results for the randomly selected label}
\label{fig:abl_mix_rand}
\end{figure}

We further explore the influence of multi-label selection and the granularity setting on the experimental results. The results are recorded in Table \ref{tab:abl-mlmg} and Figure \ref{fig:abl_mix_rand}.
EDMB no longer relies on multi-label data, there are two label processing methods for multi-label data, mixed label of RCF \cite{liu2019richer} or randomly selected label of UAED \cite{zhou2023treasure}.  
Granularity setting means the setting of hyperparameter $\gamma$ in Eq. \ref{eq-sampling}.
% Multi-granularity edges in EDMB are generated in a similar way as UAED, so UAED is used as a reference. 
% Due to the different supervision information used by EDMB and UAED, EDMB cannot achieve the best results with the same $\gamma$ of UAED.
By observing the results of different granularities in Figure \ref{fig:abl_mix_rand}, it is found that when $\gamma\in [-5,0]$, the granularity of EDMB changes most obviously, and the accuracy of multi-granularity edge in Table \ref{tab:abl-mlmg} is also the highest.
When $\gamma>0$, the granularity of the edges hardly changes anymore.
The mixed label performs worse than randomly selected label, not only less edge granularity change as shown in Figure \ref{fig:abl_mix_rand}, but also lower multi-granularity accuracy.

\section{Conclusion}
EDMB leverages Mamba to efficiently capture long-range dependencies and fine-grained cues of images, which surpasses Transformer-based detectors in terms of accuracy and efficiency. 
In addition, we introduce ELBO loss to supervise learnable Gaussian distributions that are able to generate multi-granularity edges without relying on multi-labeled data.
EDMB is the first edge detector that can generate multi-granularity edges on single-label datasets and achieves competitive results.
\textbf{Limitation.} The computational cost of EDMB is higher than CNN. And there is a lack of discussion on the application of multi-granularity edges to downstream tasks.

% We should make EDMB more lightweight. 
% The accuracy of multi-granularity test is hard to reflect the beauty of multi-granularity edge.
% We are supposed to explore the value of multi-granularity edge in more downstream applications.

% We should demonstrate the advantage of multi-granularity edges in more downstream tasks, not only in the accuracy of edge detection.

%%%%%%%%% REFERENCES
{\small
\bibliographystyle{ieee_fullname}
\bibliography{PaperForReview}
}

\end{document}